%% file: main.tex
\documentclass[twoside]{article}

\usepackage[accepted]{aistats2026}
\usepackage{amsfonts}
\usepackage{amsthm}
\usepackage{graphicx}
\usepackage{placeins}
\usepackage{booktabs}
\usepackage{xcolor}
\usepackage{hyperref}

\usepackage[round]{natbib}

\begin{document}

\twocolumn[

\runningtitle{Same Graph, Different Likelihoods}
\runningauthor{Fredsgaard, Thomas, Andersen, Schmidt, and Sugiyama}

\aistatstitle{Same Graph, Different Likelihoods: Calibration of Autoregressive Graph Generators via Permutation-Equivalent Encodings}

\aistatsauthor{
    \begin{tabular}[t]{c}\bfseries Laurits Fredsgaard\\\mdseries\small Technical University of Denmark\end{tabular} \And
    \begin{tabular}[t]{c}\bfseries Aaron Thomas\\\mdseries\small University of Birmingham\end{tabular} \And
    \begin{tabular}[t]{c}\bfseries Michael Riis Andersen\\\mdseries\small Technical University of Denmark\end{tabular} \AND
    \begin{tabular}[t]{c}\bfseries Mikkel N. Schmidt\\\mdseries\small Technical University of Denmark\end{tabular} \And
    \begin{tabular}[t]{c}\bfseries Mahito Sugiyama\\\mdseries\small National Institute of Informatics\end{tabular}
}
\aistatsaddress{}
]

\begin{abstract}
Autoregressive graph generators define likelihoods via a sequential construction process, but these likelihoods are only meaningful if they are consistent across all linearizations of the same graph.
Segmented Eulerian Neighborhood Trails (SENT), a recent linearization method, converts graphs into sequences that can be perfectly decoded and efficiently processed by language models, but admit multiple equivalent linearizations of the same graph.
We quantify violations in assigned negative log-likelihood (NLL) using the coefficient of variation across equivalent linearizations, which we call \emph{Linearization Uncertainty} (LU).
Training transformers under four linearization strategies on two datasets, we show that
biased orderings achieve lower NLL on their native order but exhibit expected calibration error (ECE) two orders of magnitude higher under random permutation, indicating that these models have learned their training linearization rather than the underlying graph.
On the molecular graph benchmark QM9, NLL for generated graphs is negatively correlated with molecular stability (AUC~$=0.43$), while LU achieves AUC~$=0.85$, suggesting that permutation-based evaluation provides a more reliable quality check for generated molecules.
Code is available at \url{https://github.com/lauritsf/linearization-uncertainty}
\end{abstract}

\section{Introduction}
Autoregressive models on permutation-invariant objects such as graphs and molecules need to linearize them into sequences before applying the chain rule of probability.
Prior work has asked which ordering gives better generation metrics~\citep{DBLP:journals/corr/VinyalsBK15, you2018graphrnn, bu2023let}.
We ask a different question: does the choice of ordering affect whether the model's likelihoods are \emph{trustworthy}?

The Segmented Eulerian Neighborhood Trail (SENT) encoding~\citep{autograph2024} makes this question testable.
Every SENT linearization of a graph decodes back to the exact same topology; the encodings are \emph{permutation-equivalent}.
Under a consistent joint distribution, the chain rule therefore implies that all orderings should yield identical negative log-likelihood (NLL).
Any observed NLL difference is therefore a model property, not an encoding artifact. This distinguishes our study from prior ordering comparisons where encoding-level differences confound the analysis.

This paper makes three contributions.
First, we formalize the linearization-invariance requirement and introduce linearization uncertainty (LU) as a metric to quantify deviations from it (Section~\ref{sec:methods}).
Second, we characterize how the linearization strategy affects calibration on both a small-graph dataset and on the standard molecular graph benchmark dataset QM9~\citep{ramakrishnan2014quantum}: While biased orderings achieve lower native NLL, expected calibration error (ECE) rises by two orders of magnitude under random permutation (Section~\ref{sec:qm9_results}).
Third, we show that Generation NLL is negatively correlated with molecular stability on QM9 (AUC~$=0.43$), while LU achieves AUC~$=0.85$ (Section~\ref{sec:nll_paradox}).

\section{Methods}
\label{sec:methods}

\subsection{SENT Encoding}
We build upon the SENT framework~\citep{autograph2024}, which linearizes graphs into sequences of tokens by sampling segmented trails that cover every edge of the full graph exactly once while incorporating neighborhood information.

\subsection{Linearization Strategies}
\label{sec:strategies}

We evaluate the effect of the linearization by comparing four distinct graph traversal strategies. All produce valid SENT encodings of the same graph (i.e., token sequences that cover every edge exactly once and decode back to the original graph topology), but they induce different inductive biases in the Transformer:

\begin{itemize}
    \item \textbf{Random Order:} At every decision point (starting node, trail extension, and jump to the next unvisited component), the candidate is selected uniformly at random; see \citet{autograph2024} for the full sampling procedure. Each training epoch therefore presents the model with a different linearization of every graph. This is the maximally diverse strategy.
    \item \textbf{Min-Degree First:} Traversal begins from a minimum-degree node (leaf). This simplifies the grammar by deferring hub nodes (which participate in many cycles) to later in the sequence.
    \item \textbf{Max-Degree First:} Traversal begins from a maximum-degree node (hub). This front-loads structural complexity.
    \item \textbf{Anchor Expansion:} Traversal begins at the maximum-degree node (the \emph{anchor}) and expands outward by preferring minimum-degree (leaf) neighbors at each step. When the trail reaches a dead end, traversal jumps to the next-highest-degree unvisited node and resumes leaf-first from there.
\end{itemize}

In all cases, ties are broken uniformly at random.

\subsection{Linearization Uncertainty and Calibration}
\label{sec:tu}

\paragraph{Formal grounding.}
Let $\phi: \mathcal{S} \to \mathcal{G}$ denote the SENT decoding map from sequences to graphs (many-to-one).
Since all sequences in the pre-image $\phi^{-1}(G)$ decode to the same graph $G\in\mathcal{G}$, any distribution over graphs induces equal likelihood across all linearizations of $G$.
For consistency across linearizations, autoregressive models $p_\theta(s) = \prod_t p_\theta(s_t \mid s_{<t})$ must therefore satisfy
$$-\log p_\theta(s) = -\log p_\theta(s') \quad \text{for all } s, s' \in \phi^{-1}(G),$$
despite the two sequences potentially having different lengths and entirely different conditional factorizations (since SENT sequence length depends on the traversal; see Appendix~\ref{app:token_analysis}).
Standard autoregressive training via teacher forcing does not enforce this constraint, so violations are expected in practice.

To quantify how far the model's sequence-level likelihoods violate this invariance requirement, we compute the coefficient of variation of the NLL across different linearizations. We refer to this quantity as linearization uncertainty (LU).

\paragraph{Linearization Uncertainty ($\mathrm{LU}$).}
Given a graph $G$, we sample $K$ sequences $\{s_1, \dots, s_K\}$ and define:
$$\mathrm{LU}(G) = \frac{\sigma(\{ \mathcal{L}(s_1), \dots, \mathcal{L}(s_K) \})}{\mu(\{ \mathcal{L}(s_1), \dots, \mathcal{L}(s_K) \})},$$
where $\mathcal{L}(s) = -\log p_\theta(s)$ and $\mu(\cdot)$ and $\sigma(\cdot)$ denote the sample mean and standard deviation, respectively.

We evaluate LU in two settings: under the model's native training strategy, measuring internal consistency, and under random permutations, measuring robustness to out-of-distribution orderings.
LU requires only scalar NLL from $K$ forward passes (no full logit access) and forward passes can be batched.

\paragraph{Calibration (ECE).}
We also measure Expected Calibration Error~\citep{guo2017calibration} per-token across the test set ($B = 15$ equal-width bins).
Full ECE decomposition by token type (new-node, revisit, node label, edge label, special) is reported in Appendix~\ref{app:cross_eval}.

If the model assigns different NLL to equivalent inputs, its distribution over graphs is not well-defined, let alone calibrated.
ECE measures whether predicted token probabilities match empirical frequencies; LU checks whether the model assigns the same likelihood to the same graph under different linearizations.
The two diagnostics are complementary.

\paragraph{Generated vs.\ Resampled Likelihoods.}
We distinguish the \emph{Generation NLL} $(-\log p_\theta(s_{\text{gen}}))$ of the sequence actually produced by the model from the \emph{mean permutation NLL} of $K$ algorithmically re-linearized versions of the same decoded graph.
A gap between these reflects specialization to the training linearization.

\section{Experiments}
\paragraph{Data.}
We evaluate our approach across two regimes: data-scarce settings using the synthetic Planar dataset~\citep{martinkus2022spectre} ($N = 128$ training graphs) as well as small subsets of QM9~\citep{ramakrishnan2014quantum}, and a data-rich setting using the full QM9 dataset.
QM9 contains ${\approx}134$k small organic molecules; following the splits and explicit-hydrogen preprocessing of \citet{autograph2024}, this yields $N \approx 98$k training molecules with 5 atom types and 4 bond types.

\paragraph{Model.}
Following \citet{autograph2024}, we employ a 12-layer Llama Transformer backbone and use constrained decoding at inference time on QM9. This sets the logits to $-\infty$ for any tokens that would result in an invalid SENT grammar, ensuring the model always outputs a decodable graph topology without explicitly enforcing chemical validity.

\paragraph{Metrics.} Our evaluation considers both general probabilistic metrics (such as NLL, ECE and LU), and domain-specific generative metrics. The latter comprise Validity (dataset-specific structural checks: planarity for Planar and RDKit sanitization for QM9), Uniqueness (fraction of non-duplicates), and Novelty (fraction of graphs absent from the training set). For QM9, we additionally report chemical stability (valency compliance), Fr\'{e}chet ChemNet Distance (FCD), and PolyGraph Discrepancy (PGD). Detailed definitions are provided in Appendix~\ref{app:metric_defs}.

\subsection{Data-scarce regime.}
On Planar ($N = 128$), under the biased (non-random) linearization strategies the model quickly memorizes the training sequences: validation NLL rises after ${\sim}5$k steps. While overall generative quality appears to improve, a component breakdown reveals this is driven entirely by Validity. Uniqueness and Novelty for biased strategies gradually worsen to 90--95\% and 75--85\%, respectively, as the models increasingly reproduce training graphs (Appendices~\ref{app:planar}, \ref{app:planar_vun_components}). This memorization effect persists at larger scales: On QM9 subsets, biased strategies only recover competitive Uniqueness once $N$ reaches $10{,}000$, and their sequence diversity saturates early across all dataset scales (Appendices~\ref{app:subset_sweep}, \ref{app:diversity_saturation}). In contrast, the large number of permutations available under Random Order act as inherent data augmentation, preventing overfitting and sustaining both high diversity and novelty.

\subsection{Scaling to Data-Rich Regimes}
\label{sec:qm9_results}

\input{tables/permutation_consistency_qm9.tex}

In the data-rich regime (QM9 Full, $N \approx 98$k), training stabilizes across all strategies.
Biased strategies achieve lower native NLL/token and shorter sequences, reflecting traversal biases that minimize chordal back-references.
Overall generative quality remains comparable across strategies (Table~\ref{tab:generation-quality} in Appendix~\ref{app:gen_quality_qm9}).

\paragraph{The robustness trade-off.}
Biased models act as better density estimators under the specific linearizations they were trained on (native evaluation): Anchor Expansion achieves the lowest NLL/tok and LU (Table~\ref{tab:robustness}).
Under randomized evaluation, however, NLL/tok for biased strategies rises by up to $10{\times}$ and LU increases by an order of magnitude (``Random'' columns in the table).

Table~\ref{tab:robustness} shows that this structural brittleness co-occurs with a collapse in absolute probability calibration: ECE for biased strategies rises by two orders of magnitude off-distribution.
This shows that these models have learned their training linearization rather than the underlying graph topology.
Random Order does not degrade, since its native strategy already samples uniformly.
The full $4{\times}4$ cross-evaluation matrices (Appendix~\ref{app:cross_eval}) show that Revisit tokens (cycle-closing back-references) account for the largest ECE failure off-diagonal, while New Node tokens remain well-calibrated across orderings.

\input{tables/generation_self_assess_qm9.tex}

The previous analysis evaluates the model on held-out test graphs.
We now ask whether it can reliably score its own generations.

\paragraph{Self-assessment of generated sequences.}
Table~\ref{tab:generation-self-assess} shows that mean permutation NLL is consistently higher than Generation NLL across all strategies.
Generated sequences are also longer than algorithmically-resampled sequences of the same molecules (e.g.\ $127$ vs.\ $89$ tokens for Random Order). This length gap occurs because the model learns valid but structurally inefficient traversals, such as deferring explicit hydrogens, which incur nearly twice as many chordal back-references as the greedy algorithmic linearizer (Appendix~\ref{app:token_analysis}).

\subsection{Linearization Uncertainty as a Molecule Quality Signal}
\label{sec:nll_paradox}

Under constrained decoding, molecules that fail chemical stability checks are \emph{more} likely (have lower Generation NLL) than stable ones (AUC~$=0.43$, averaged across strategies and seeds). A likely explanation is that unstable molecules are generated through traversals that happen to align well with the training linearization: each token prediction is confident, producing low NLL, but the resulting global structure violates valency.
When the same graph is re-linearized under different orderings, the model encounters unfamiliar token sequences and NLL rises.
The variance across orderings is therefore larger for molecules whose low Generation NLL depended on a specific traversal, rather than reflecting genuine graph-level confidence.

Per-molecule LU ($K{=}32$) achieves AUC~$=0.85$ for binary stability prediction, compared to AUC~$=0.43$ for Generation NLL (Figure~\ref{fig:stability_predictors}).
ECE achieves AUC~$=0.89$ but requires full logit access; LU achieves comparable predictive signal from scalar NLL alone.
As shown in Figure~\ref{fig:k_sweep}, this performance is highly sample efficient (more details in Appendix~\ref{app:stability_predictors}).

\begin{figure}[!ht]
    \centering
    \includegraphics[width=\linewidth]{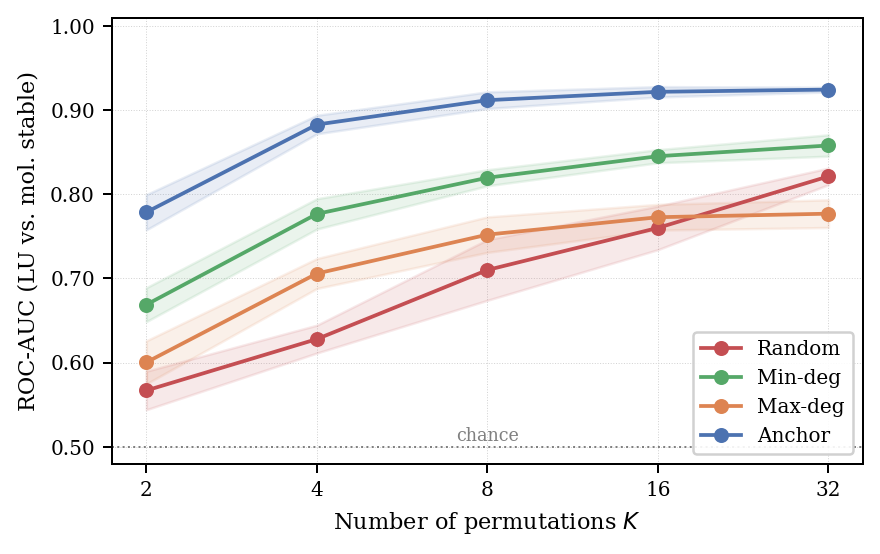}
    \caption{\textbf{LU AUC vs.\ number of permutations $K$ (QM9).} ROC-AUC for predicting molecular stability from LU, per linearization strategy (mean $\pm$ std across 3 seeds). Even at $K{=}2$ the signal far exceeds the Generation NLL baseline (AUC~$=0.43$).}
    \label{fig:k_sweep}
\end{figure}

\section{Conclusion and Discussion}

Because SENT encodings are permutation-equivalent, any non-zero CV of NLL across orderings is a violation of graph-consistency, not a consequence of encoding choice.
LU needs only scalar NLL from $K$ forward passes and does not require full logit access.

Across both datasets, biased strategies converge to their training linearization rather than the underlying graph structure.
On QM9 this produces lower native NLL but ECE two orders of magnitude higher under random permutation.
Generation NLL should not be used alone as a quality filter: it is negatively correlated with molecular stability (AUC~$=0.43$), while LU correctly identifies unstable molecules (AUC~$=0.85$).
Training with random linearization is a straightforward intervention to obtain permutation-consistent likelihoods, and permutation-based evaluation should be used as a secondary check when generating structured objects.

The stability predictor results are limited to QM9, which is a constrained small-molecule dataset; larger and more chemically diverse benchmarks such as MOSES or GuacaMol would better test whether the Generation NLL inversion and the LU signal hold more broadly.
The analysis is also limited to the SENT framework.
$K$-sensitivity is characterized in Appendix~\ref{app:stability_predictors}; whether LU converges at similar $K$ on larger graphs is an open question.

\subsubsection*{Acknowledgements}
This work was conducted while L.\ Fredsgaard and A.\ Thomas were visiting researchers at the National Institute of Informatics, Tokyo.
We thank Dexiong Chen for meaningful discussions about graph linearization and the SENT algorithm.
This work was supported by JST, CREST Grant Number JPMJCR22D3, Japan.
The authors acknowledge support from the Novo Nordisk Foundation under grant no NNF22OC0076658 (Bayesian neural networks for molecular discovery).

\bibliography{references}

\clearpage
\appendix
\thispagestyle{empty}

\onecolumn
\aistatstitle{Same Graph, Different Likelihoods: \\
Supplementary Materials}

\section{Metric Definitions}
\label{app:metric_defs}

\FloatBarrier

All percentage metrics (Validity, Uniqueness, Novelty, Atm.\ Stable, Mol.\ Stable) are reported as fractions of the generated set.
QM9 metrics in Table~\ref{tab:generation-quality} are computed on 10{,}000 generated molecules per model.

\paragraph{Probabilistic and Robustness Metrics.}
\begin{description}
    \item[NLL] Negative Log-Likelihood ($-\log p_\theta(s)$). We distinguish between the \emph{Generation NLL} of the model's own output and the \emph{Mean Permutation NLL} averaged over $K$ re-linearizations of the same graph.
    \item[LU] \textbf{Linearization Uncertainty}. The coefficient of variation ($\sigma/\mu$) of the NLL across $K$ equivalent linearizations of the same graph. It quantifies a model's deviation from linearization-invariance.
    \item[ECE] Expected Calibration Error \citep{guo2017calibration}. Measures the correspondence between predicted token probabilities and empirical accuracy, computed across $B=15$ equal-width bins.
\end{description}

\paragraph{Metrics common to both datasets.}
\begin{description}
    \item[Validity] Fraction of generated outputs that satisfy dataset-specific structural constraints. For QM9: the decoded token sequence yields a valid SMILES string (RDKit \texttt{MolFromSmiles} with sanitization). For Planar: the decoded graph passes a planarity check.
    \item[Uniqueness] Fraction of valid generated outputs that are distinct (by canonical SMILES for QM9, by graph isomorphism for Planar).
    \item[Novelty] Fraction of unique valid outputs absent from the training set.
    \item[VUN] Validity $\times$ Uniqueness $\times$ Novelty. Used as a composite generative quality score in the Planar experiments (Appendix~\ref{app:planar}).
\end{description}

\paragraph{QM9-specific metrics.}
\begin{description}
    \item[Atm.\ Stable] Fraction of generated atoms satisfying strict valency constraints (H:1, C:4, N:3, O:2, F:1, B:3, Si:4, P:3/5, S:4, Cl:1, Br:1, I:1; following \citet{hoogeboom2022equivariant}).
    \item[Mol.\ Stable] Fraction of molecules in which every atom satisfies these constraints simultaneously.
    \item[FCD] Fr\'{e}chet ChemNet Distance~\citep{preuer2018frechet}: Fr\'{e}chet distance between 128-dimensional ChemNet embeddings of the test set and generated molecules; lower is better.
    \item[PGD] PolyGraph Discrepancy~\citep{krimmel2025polygraph}: estimated Jensen--Shannon distance between real and generated graph distributions, obtained by training a binary classifier on a descriptor suite including RDKit chemoinformatics features (topological indices, Lipinski descriptors, Morgan fingerprints), 128-dimensional ChemNet LSTM embeddings, and MolCLR contrastive GNN representations~\citep{molclr}; only chemically valid generated molecules are evaluated; lower is better.
\end{description}

\clearpage
\section{Generation Quality on QM9}
\label{app:gen_quality_qm9}

Table~\ref{tab:generation-quality} compares all strategies on the full QM9 dataset on standard molecular generation metrics.

Note that while biased strategies exhibit lower Novelty compared to Random Order, we do not consider this a degradation in generative quality. As noted by prior work \citep{vignac2023digress, vignac2022topn}, QM9 is essentially an exhaustive enumeration of small molecules satisfying specific structural constraints. Because of this, generating "novel" molecules outside of this exhaustive set often indicates a failure to capture the strict underlying data distribution rather than a superior generative capability.
Conversely, these biased strategies achieve better (lower) FCD scores.

\input{tables/generation_quality_qm9.tex}

\clearpage
\section{Data-Scarce Regime: Planar Experiments}
\label{app:planar}

\begin{figure}[!ht]
    \centering
    \includegraphics[width=\linewidth]{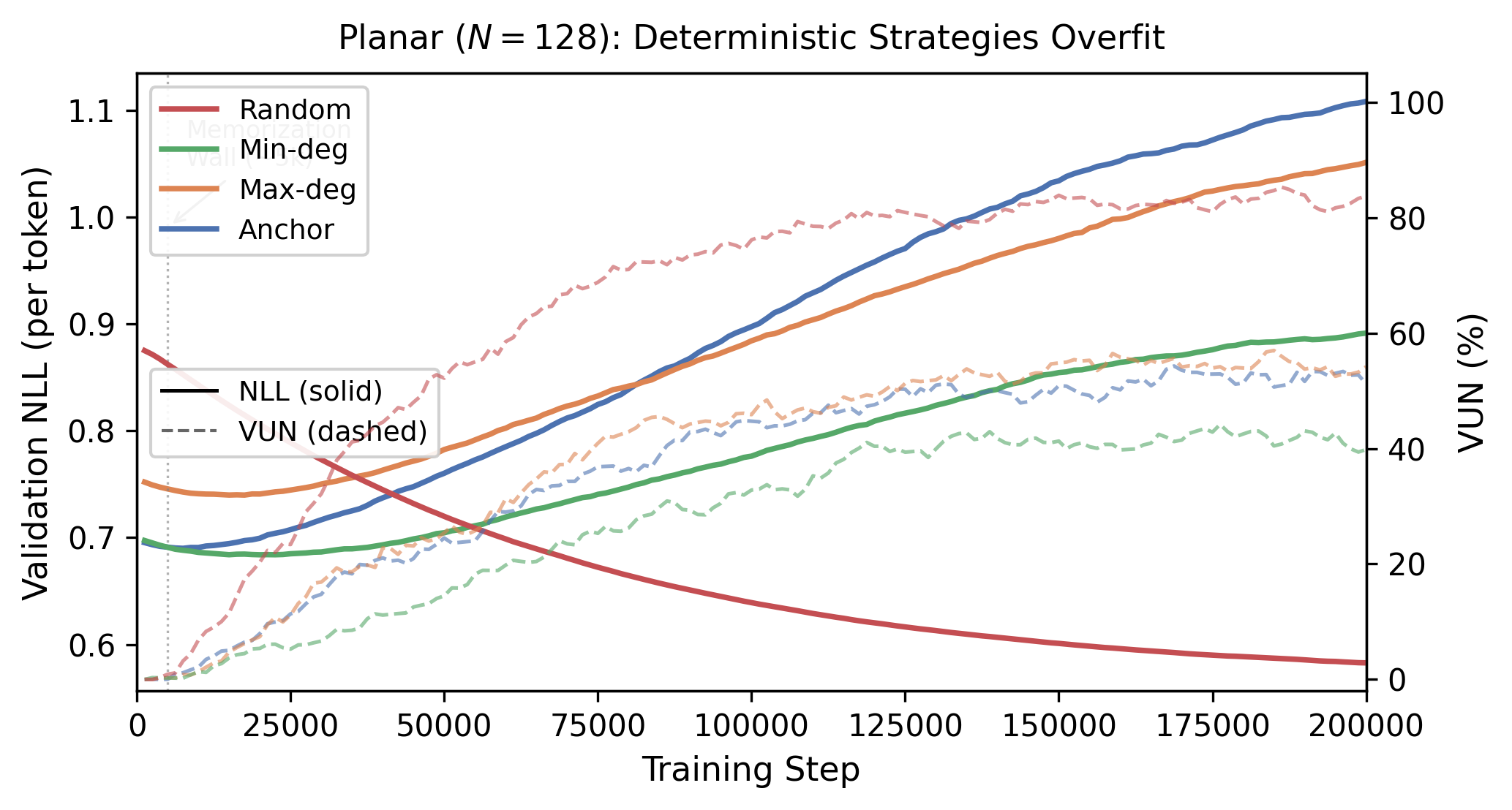}
    \caption{\textbf{Overfitting vs.\ Generative Quality in Data-Scarce Regimes.} Dual-axis plot showing validation NLL/token (solid, left axis) and VUN (dashed, right axis) for Planar ($N{=}128$, seed 0). For all biased strategies, NLL rises after $\sim$5k steps (memorization) while VUN continues to climb, driven by Validity alone. Only Random Order achieves sustained improvement in both axes. See Appendix~\ref{app:planar_vun_components} for the decomposition of VUN into Validity, Uniqueness, and Novelty.}
    \label{fig:planar_overfitting}
\end{figure}

\clearpage
\section{From Memorization to Generalization: A Regime Analysis}
\label{app:qm9_reduced}

\subsection{Performance Sensitivity to Dataset Size}
\label{app:subset_sweep}

\begin{figure}[!ht]
    \centering
    \includegraphics[width=\linewidth]{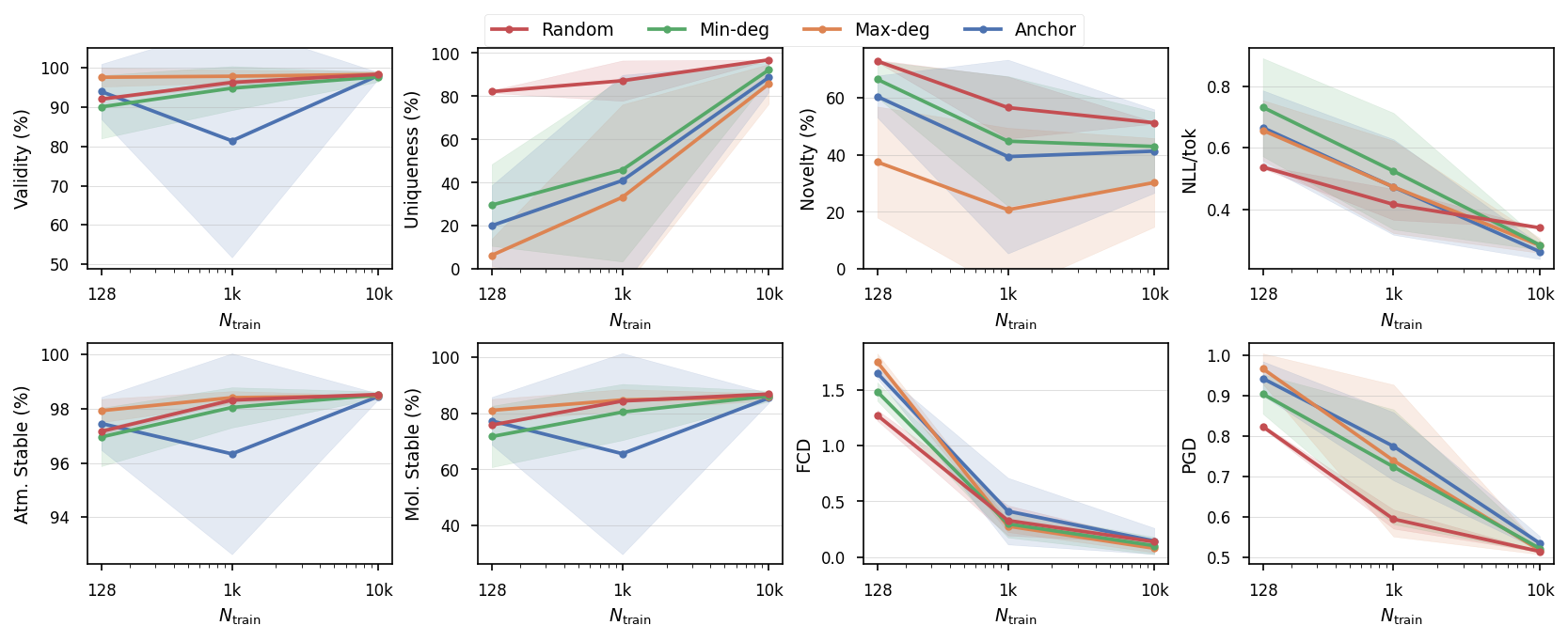}
    \caption{\textbf{QM9 subset-size sweep (mean $\pm$ std, 3 seeds).} Generative quality
across $N_{\mathrm{train}} \in \{128, 1000, 10000\}$. Uniqueness collapses for
biased strategies at small $N$, recovering only at $N{=}10{,}000$. Novelty remains
consistently lower for biased strategies at all sizes.}
    \label{fig:subset_sweep}
\end{figure}

\subsection{VUN Component Curves for Planar}
\label{app:planar_vun_components}

\begin{figure}[!ht]
    \centering
    \includegraphics[width=\linewidth]{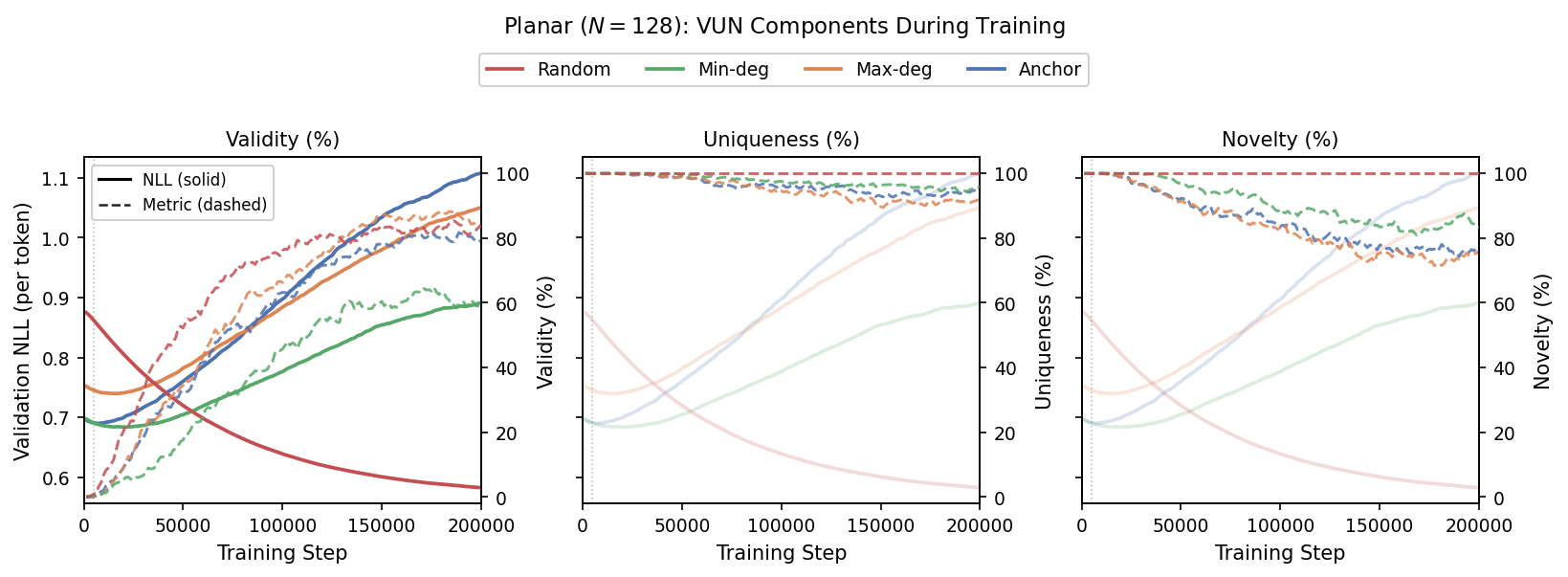}
    \caption{\textbf{Decomposition of VUN into Validity, Uniqueness, and Novelty (Planar, $N{=}128$, seed 0).} Each panel shares the NLL curve (faded solid lines, left axis) with the main figure. \emph{Validity} (left): Min-Degree First is the only strategy where Validity visibly suffers, indicating incomplete grammar acquisition. \emph{Uniqueness} (center): all biased strategies generate repeated outputs over time, with Anchor Expansion degrading most. \emph{Novelty} (right): biased strategies reproduce training examples, with Anchor Expansion collapsing to $\sim$60\%. Random Order sustains 100\% across all three components throughout training.}
    \label{fig:planar_vun_components}
\end{figure}

\subsection{Diversity Saturation Analysis}
\label{app:diversity_saturation}

\begin{figure}[!ht]
    \centering
    \includegraphics[width=\linewidth]{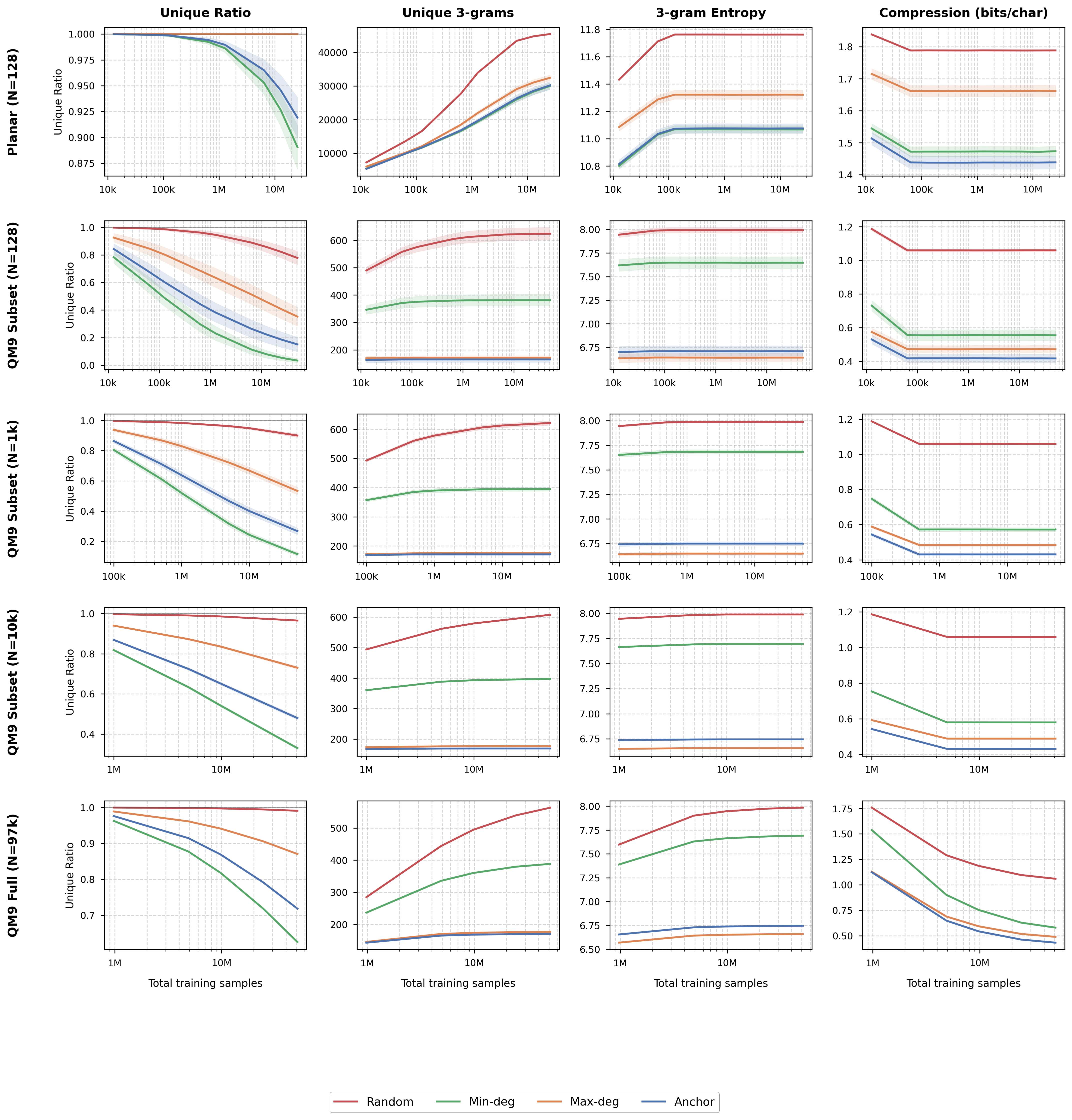}
    \caption{\textbf{Full Diversity Saturation Grid.} Diversity metrics across Planar (top),
QM9 Subset (middle), and QM9 Full (bottom). Random Order maintains higher sequence
diversity across all scales; biased strategies saturate early.}
    \label{fig:full_diversity_grid}
\end{figure}

\clearpage
\subsection{Cross-Strategy Evaluation: Full Breakdown}
\label{app:cross_eval}

\begin{figure}[!ht]
    \centering
    \includegraphics[width=\linewidth]{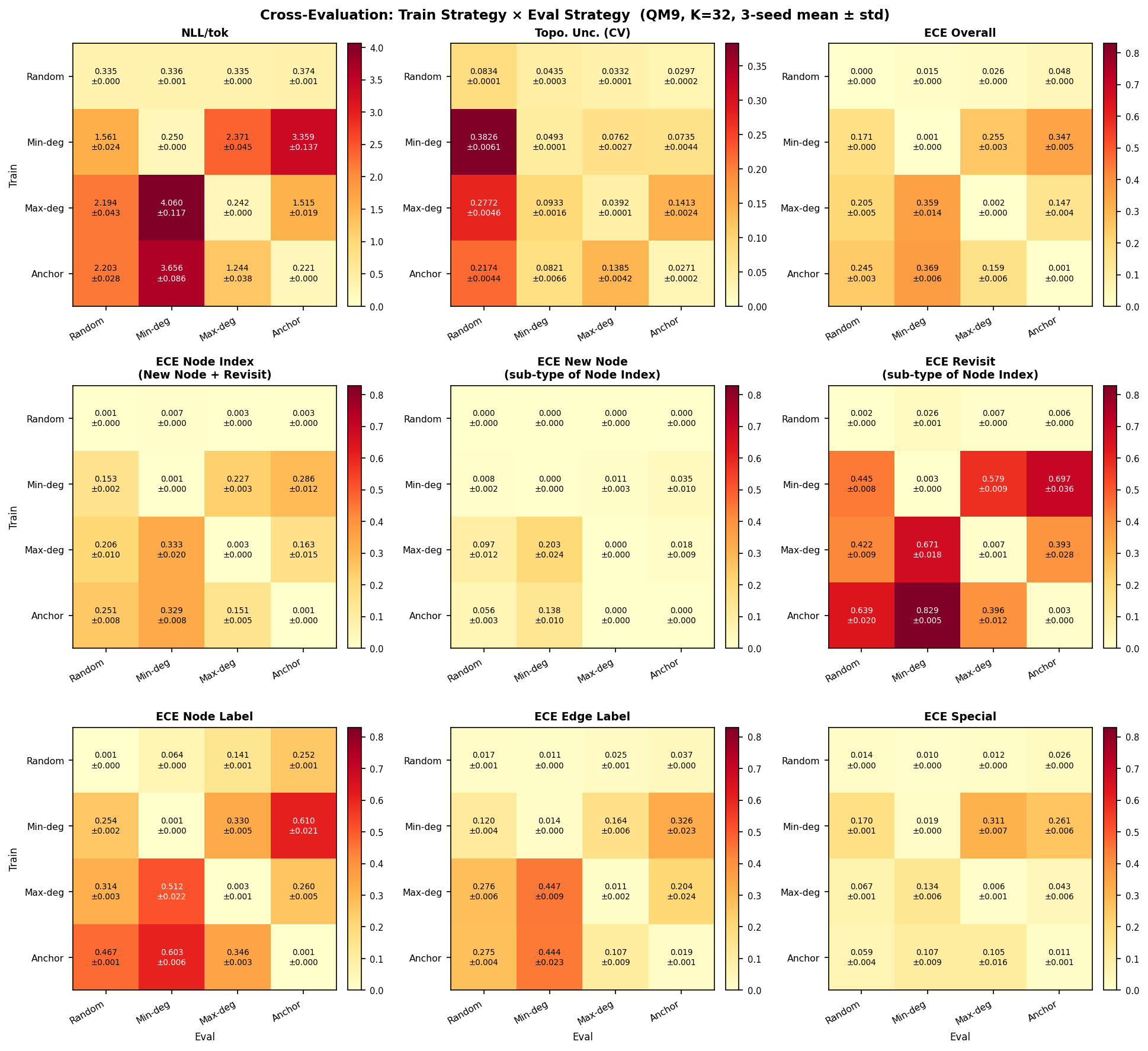}
    \caption{%
        \textbf{Full cross-evaluation breakdown (QM9, $K{=}32$, 3-seed mean $\pm$ std).}
        Rows = training strategy; columns = evaluation strategy.
        Diagonal cells (native evaluation) are well-calibrated and low-NLL for all strategies.
        Off-diagonal cells expose the ``specialist'' brittleness of biased strategies: NLL/tok, linearization uncertainty, and ECE all increase substantially when the evaluation ordering differs from the training ordering.
        Random order (top row/column) is the most robust across all metrics.
        Among ECE token types, \textbf{Revisit} tokens show the largest absolute calibration failure for biased strategies evaluated off-diagonal, confirming that these models exploit topological sorting cues to resolve cycle-closing decisions.
    }
    \label{fig:cross_eval_heatmaps}
\end{figure}

Figure~\ref{fig:cross_eval_heatmaps} shows all $4 \times 4$ cross-evaluation results (train strategy $\times$ eval strategy) for nine metrics: NLL per token, linearization uncertainty (LU), overall ECE, and ECE decomposed by the six token types (Node Index, Node Label, Edge Label, Special, New Node, Revisit).
Each cell reports the mean $\pm$ std across three seeds on the full QM9 test set ($K{=}32$).

The diagonal entries confirm that every strategy is well-calibrated and achieves low NLL under its own native ordering.
Off-diagonal entries tell a different story: biased strategies (Min-Degree, Max-Degree, Anchor) suffer dramatic increases in both NLL/tok (up to $\sim$4$\times$) and ECE when evaluated under a foreign ordering, whereas Random remains relatively stable across all four eval columns (NLL/tok range 0.335--0.374).

Decomposing ECE by token type reveals the mechanism of failure.
We distinguish two sub-types within Node Index predictions: \textbf{New Node} (the first occurrence of a node index, a grammatical prediction) and \textbf{Revisit} (back-references to form cycles, a topological prediction).
When biased models are evaluated on random linearizations, New Node ECE remains low (${\leq}0.01$ for Min-Degree), showing that the grammar of node addition is learned robustly.
Revisit ECE shows the most extreme off-diagonal failure ($\approx$0.44 for Min-Degree, $\approx$0.64 for Anchor): without the native topological sorting, the model cannot identify which node to connect back to.
Node Label and Edge Label ECE also rise substantially off-diagonal; only New Node tokens remain well-calibrated across orderings.
The calibration degradation is therefore not confined to revisit tokens but manifests across most token types that depend on the relative position of nodes in the sequence.
Random order also loses some calibration when evaluated on structured orderings, though the degradation is far milder than in the reverse direction (NLL/tok range 0.335--0.374).

\clearpage
\section{Stability Predictor Analysis}
\label{app:stability_predictors}

\begin{figure}[!ht]
    \centering
    \includegraphics[width=\linewidth]{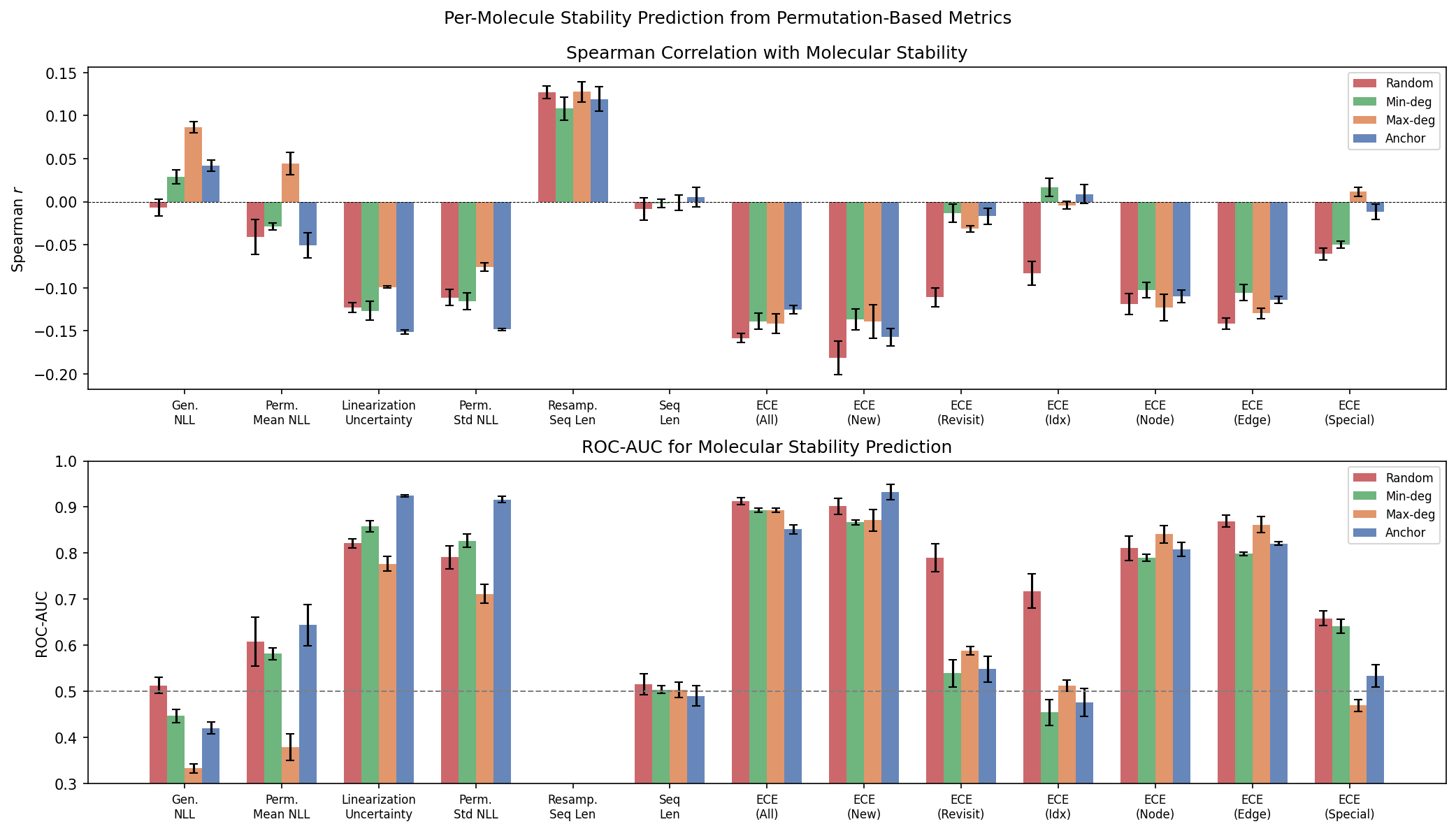}
    \caption{\textbf{Per-molecule stability prediction from permutation-based metrics (QM9).} Each bar shows mean $\pm$ std across 4 strategies $\times$ 3 seeds. \emph{Top}: Spearman correlation with molecular stability. \emph{Bottom}: ROC-AUC for binary stability prediction. Generation NLL is negatively correlated with stability; LU achieves AUC~$=0.85$.}
    \label{fig:stability_predictors}
\end{figure}

Figure~\ref{fig:stability_predictors} reports the predictive power of per-molecule permutation-based metrics for chemical stability on QM9, averaged across four strategies and three seeds.
To compute the ROC-AUC without training a classifier, we use molecular stability (valency compliance) as the positive binary label and the raw metric as the continuous prediction score. We negate the metric values prior to computation so that lower scores predict the positive class.
Consequently, an AUC of 0.85 indicates an 85\% probability that a randomly selected stable molecule has a lower LU than an unstable one. In contrast, Generation NLL yields an AUC of 0.43, meaning the model typically assigns lower NLL (higher confidence) to unstable molecules.

Mean permutation NLL (average NLL over $K{=}32$ random re-linearizations) is close to chance (AUC~$=0.55$).
Linearization Uncertainty (CV of NLL across $K{=}32$ permutations) achieves AUC~$=0.85$.
ECE achieves AUC~$=0.89$.

\paragraph{K-sensitivity.}
Figure~\ref{fig:k_sweep} (main text) shows a full sweep over $K \in \{2, 4, 8, 16, 32\}$, subsampled from a single 32-permutation evaluation run (so all $K$ values share the same permutation draws and are directly comparable).
Overall mean AUC rises from $0.65$ at $K{=}2$ to $0.85$ at $K{=}32$, but convergence speed varies by strategy.
Anchor Expansion reaches $0.90$ already at $K{=}8$ and gains only $0.02$ further by $K{=}32$, while Random Order is still climbing steeply at $K{=}32$ ($0.82$), suggesting that models with no traversal bias require more permutations to produce a stable LU estimate.
Across all $K$, the qualitative ranking is preserved: even at $K{=}2$, LU (AUC~$=0.65$) far outperforms Generation NLL (AUC~$=0.43$).

\clearpage
\section{Sequence Length Gap: Mechanistic Analysis}
\label{app:token_analysis}

The sequence length of a SENT encoding decomposes into a fixed \emph{graph invariant} (determined entirely by the number of atoms and bonds) and a traversal-dependent \emph{navigational overhead}. This overhead increases when a traversal is fragmented into more individual train segments, which requires additional non-empty neighborhood sets and chordal back-references to close rings and connect the graph.
Because generated and test molecules have comparable atom and bond counts, the observed length gap stems directly from differences in this navigational overhead. Model-generated sequences are less efficient, incurring more segments and chordal groups.
This behavior is a property of the traversal rather than the underlying molecule. The standard SENT algorithm greedily extends trails which minimizes back-references. The autoregressive model, however, learns no such strict constraint and often produces implicit orderings that are valid but inefficient. For instance, a model might defer an entire class of atoms (such as explicit hydrogens) to late in the sequence, forcing each to require a separate back-reference to an already-visited parent atom.

\begin{figure}[!ht]
    \centering
    \includegraphics[width=\linewidth]{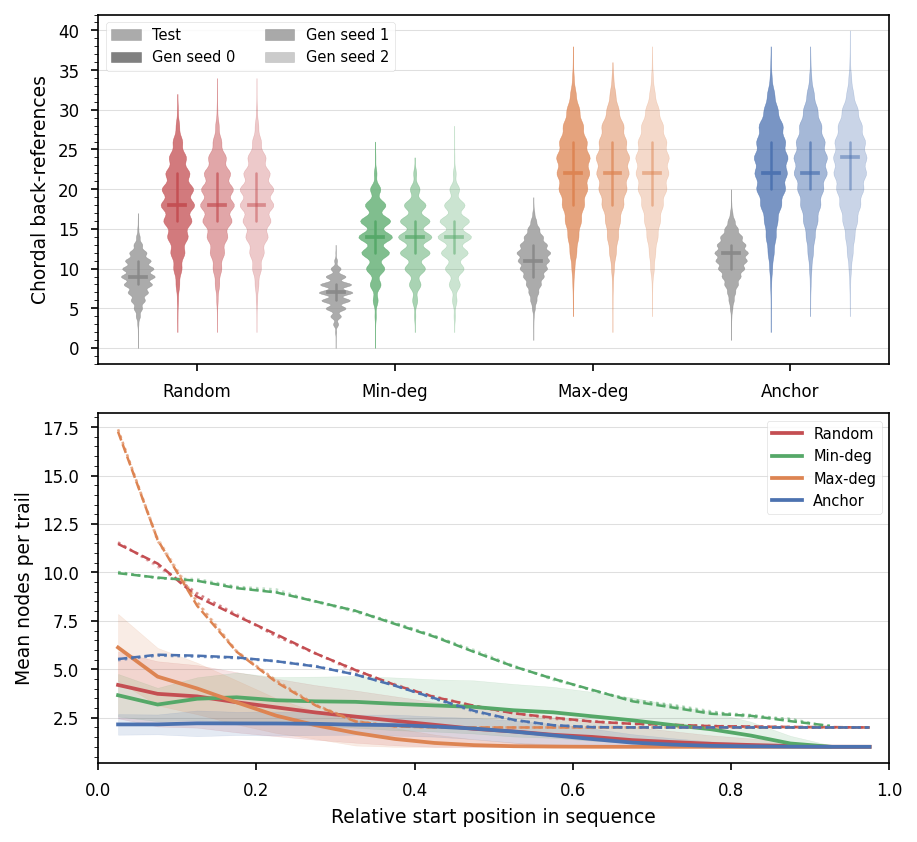}
    \caption{\textbf{Mechanistic Analysis of Sequence Length.} \emph{Top:} Distribution of chordal back-references per molecule for test graphs and model-generated sequences (across 3 seeds). Despite similar node and edge counts, model-generated sequences incur $\approx 2\times$ more back-references, explaining the length gap in Table~\ref{tab:generation-self-assess}. \emph{Bottom:} Mean trail-length profile as a function of relative sequence position. Solid lines = algorithmically-resampled test sequences per strategy; dashed lines = generated sequences. Generated sequences consistently front-load large trails compared to algorithmically sampled linearizations.}
    \label{fig:token_analysis}
\end{figure}

Figure~\ref{fig:token_analysis} illustrates this effect. The top panel shows that model-generated sequences contain roughly twice as many chordal back-references as algorithmically-linearized test graphs of the same molecules. The bottom panel displays the trail-length profile (mean nodes per trail segment across relative sequence position). While algorithmically-resampled test sequences (solid lines) exhibit strategy-dependent profiles, generated sequences (dashed lines) consistently front-load large trails regardless of the training strategy. For generated sequences, the early-sequence peak is higher than for the corresponding resampled sequences, followed by much shorter trailing segments later.

\begin{figure}[!ht]
    \centering
    \includegraphics[width=\linewidth]{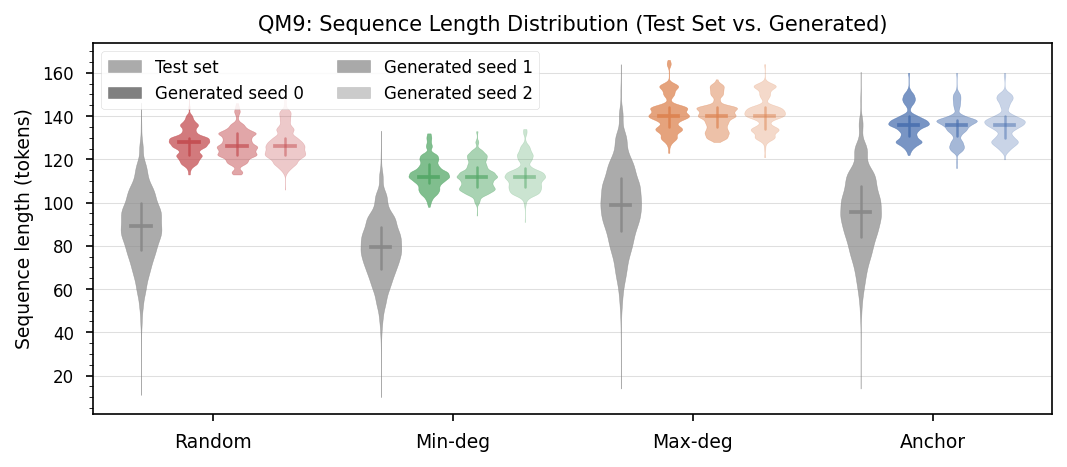}
    \caption{\textbf{QM9: Sequence Length Distribution (Test Set vs.\ Generated).} Distribution of sequence lengths for algorithmically-linearized QM9 test graphs and model-generated sequences. Generated sequences are systematically longer across all strategies.}
    \label{fig:seqlen_inset}
\end{figure}

\end{document}

%% file: tables/permutation_consistency_qm9.tex
\begin{table*}[!ht]
  \centering
  \caption{Test-set robustness on QM9 ($K{=}32$ permutations, full test set, 3 seeds). Native = model's own training strategy; Random = evaluated under the Random Order strategy.}
  \label{tab:robustness}
  \begin{tabular}{lrrrrrrr}
    \toprule
    \multicolumn{1}{l}{} & \multicolumn{1}{l}{} & \multicolumn{2}{c}{NLL/token ↓} & \multicolumn{2}{c}{Linearization Uncertainty ↓} & \multicolumn{2}{c}{ECE ↓} \\
    \cmidrule(lr){3-4} \cmidrule(lr){5-6} \cmidrule(lr){7-8}
    Strategy & Tok/graph & Native & Random & Native & Random & Native & Random \\
    \midrule
    Random & $88.9$ & $0.336_{\pm 0.001}$ & $0.336_{\pm 0.001}$ & $0.083_{\pm 0.000}$ & $0.083_{\pm 0.000}$ & $0.000_{\pm 0.000}$ & $0.000_{\pm 0.000}$ \\
    Min-Degree & $79.0$ & $0.250_{\pm 0.000}$ & $1.564_{\pm 0.028}$ & $0.049_{\pm 0.000}$ & $0.383_{\pm 0.007}$ & $0.001_{\pm 0.000}$ & $0.171_{\pm 0.000}$ \\
    Max-Degree & $98.7$ & $0.244_{\pm 0.000}$ & $2.240_{\pm 0.057}$ & $0.039_{\pm 0.000}$ & $0.277_{\pm 0.006}$ & $0.002_{\pm 0.000}$ & $0.205_{\pm 0.006}$ \\
    Anchor & $95.6$ & $0.222_{\pm 0.000}$ & $2.246_{\pm 0.033}$ & $0.027_{\pm 0.000}$ & $0.217_{\pm 0.005}$ & $0.001_{\pm 0.000}$ & $0.245_{\pm 0.004}$ \\
    \bottomrule
  \end{tabular}
\end{table*}

%% file: tables/generation_self_assess_qm9.tex
\begin{table*}[!ht]
  \centering
  \caption{Model self-assessment of generated sequences on QM9 (3-seed mean$_{\pm\text{std}}$, valid molecules only). Gen.\ = model's own generated trajectory; Resamp.\ = same molecule re-linearized via the native strategy ($K{=}32$).}
  \label{tab:generation-self-assess}
  \begin{tabular}{lrrrrr}
    \toprule
    \multicolumn{1}{l}{} & \multicolumn{2}{c}{Tok/graph} & \multicolumn{2}{c}{NLL ↓} & \multicolumn{1}{l}{} \\
    \cmidrule(lr){2-3} \cmidrule(lr){4-5}
    Strategy & (Gen.) & (Resamp.) & (Gen.) & (Resamp.) & Lin. Unc. ↓ \\
    \midrule
    Random & $127.1_{\pm 0.4}$ & $89.3_{\pm 0.2}$ & $29.383_{\pm 0.007}$ & $30.361_{\pm 0.049}$ & $0.085_{\pm 0.000}$ \\
    Min-Degree & $112.9_{\pm 0.5}$ & $79.2_{\pm 0.2}$ & $19.162_{\pm 0.035}$ & $19.781_{\pm 0.026}$ & $0.055_{\pm 0.000}$ \\
    Max-Degree & $140.9_{\pm 0.3}$ & $98.7_{\pm 0.2}$ & $23.458_{\pm 0.042}$ & $23.839_{\pm 0.052}$ & $0.041_{\pm 0.000}$ \\
    Anchor & $136.2_{\pm 0.3}$ & $95.4_{\pm 0.2}$ & $20.682_{\pm 0.055}$ & $21.166_{\pm 0.037}$ & $0.031_{\pm 0.000}$ \\
    \bottomrule
  \end{tabular}
\end{table*}

%% file: tables/generation_quality_qm9.tex
\begin{table*}[!ht]
  \centering
  \caption{Generative quality on QM9 (3-seed mean$_{\pm\text{std}}$; 10{,}000 generated molecules per model). Metric definitions in Appendix~\ref{app:metric_defs}.}
  \label{tab:generation-quality}
  \begin{tabular}{lrrrrrrr}
    \toprule
    Strategy & Validity ↑ & Unique ↑ & Novelty ↑* & Atm. Stable ↑ & Mol. Stable ↑ & FCD ↓ & PGD ↓ \\
    \midrule
    Random & $98.6_{\pm 0.1}$ & $97.3_{\pm 0.2}$ & $46.0_{\pm 1.2}$ & $98.6_{\pm 0.0}$ & $87.2_{\pm 0.3}$ & $0.067_{\pm 0.002}$ & $0.515_{\pm 0.007}$ \\
    Min-Degree & $98.6_{\pm 0.0}$ & $96.9_{\pm 0.2}$ & $37.0_{\pm 0.9}$ & $98.5_{\pm 0.0}$ & $87.2_{\pm 0.1}$ & $0.042_{\pm 0.002}$ & $0.513_{\pm 0.002}$ \\
    Max-Degree & $98.8_{\pm 0.0}$ & $96.7_{\pm 0.1}$ & $29.9_{\pm 1.0}$ & $98.5_{\pm 0.0}$ & $87.0_{\pm 0.2}$ & $0.040_{\pm 0.001}$ & $0.505_{\pm 0.002}$ \\
    Anchor & $98.7_{\pm 0.2}$ & $96.8_{\pm 0.2}$ & $36.9_{\pm 0.6}$ & $98.5_{\pm 0.0}$ & $86.8_{\pm 0.3}$ & $0.049_{\pm 0.002}$ & $0.506_{\pm 0.010}$ \\
    \bottomrule
  \end{tabular}
\end{table*}